\documentclass[10pt,twocolumn,letterpaper]{article}

\usepackage[pagenumbers]{cvpr} % To force page numbers, e.g. for an arXiv version

\usepackage[dvipsnames]{xcolor}

\pdfoutput=1

\usepackage[utf8]{inputenc} % allow utf-8 input
\usepackage[T1]{fontenc}    % use 8-bit T1 fonts
\usepackage{url}            % simple URL typesetting
\usepackage{booktabs}       % professional-quality tables
\usepackage{amsfonts}       % blackboard math symbols
\usepackage{nicefrac}       % compact symbols for 1/2, etc.
\usepackage{microtype}      % microtypography
\usepackage{xcolor}         % colors

\usepackage{times}
\usepackage{epsfig}
\usepackage{graphicx}
\usepackage{amsmath}
\usepackage{amssymb}
\usepackage{booktabs}
\usepackage{multirow}
\usepackage{gensymb}
\usepackage{bbding}
\usepackage{soul}
\usepackage{multirow}
\usepackage{array}
\usepackage{enumitem}
\usepackage{pifont}
\usepackage{graphicx}
\usepackage{subcaption}
\usepackage[toc,title,page]{appendix}

\makeatletter
\DeclareRobustCommand\onedot{\futurelet\@let@token\@onedot}
\def\@onedot{\ifx\@let@token.\else.\null\fi\xspace}

\def\wrt{\emph{w.r.t}\onedot}

\DeclareMathOperator{\sine}{sine}
\DeclareMathOperator{\norm}{norm}
\DeclareMathOperator{\Att}{Att}

\definecolor{cvprblue}{rgb}{0.21,0.49,0.74}
\usepackage[pagebackref,breaklinks,colorlinks,citecolor=cvprblue]{hyperref}

\usepackage[capitalize]{cleveref}
\crefname{section}{Sec.}{Secs.}
\Crefname{section}{Section}{Sections}
\Crefname{table}{Table}{Tables}
\crefname{table}{Tab.}{Tabs.}

\def\paperID{5219} % *** Enter the Paper ID here
\def\confName{CVPR}
\def\confYear{2024}

\title{DGC-GNN: Leveraging Geometry and Color Cues for Visual Descriptor-Free 2D-3D Matching}

\author{
Shuzhe Wang$^*$ \ \ \ \ \ \ \  \ \ \  Juho Kannala$^*$  \ \ \ \ \ \ \ \ \ \ Daniel Barath$^\dagger$\\
$^*$ Department of Computer Science, Aalto University \\
$^\dagger$ Computer Vision and Geometry Group, ETH Zurich \\
{\tt shuzhe.wang@aalto.fi \  juho.kannala@aalto.fi \ dbarath@inf.ethz.ch}
}

\begin{document}

\maketitle

\begin{abstract}
Matching 2D keypoints in an image to a sparse 3D point cloud of the scene without requiring visual descriptors has garnered increased interest due to its low memory requirements, inherent privacy preservation, and reduced need for expensive 3D model maintenance compared to visual descriptor-based methods. 
However, existing algorithms often compromise on performance, resulting in a significant deterioration compared to their descriptor-based counterparts. 
In this paper, we introduce DGC-GNN, a novel algorithm that employs a global-to-local Graph Neural Network (GNN) that progressively exploits geometric and color cues to represent keypoints, thereby improving matching accuracy. 
Our procedure encodes both Euclidean and angular relations at a coarse level, forming the geometric embedding to guide the point matching.
We evaluate DGC-GNN on both indoor and outdoor datasets, demonstrating that it not only doubles the accuracy of the state-of-the-art visual descriptor-free algorithm but also substantially narrows the performance gap between descriptor-based and descriptor-free methods.\footnote[1]{The code and trained models are available at: \url{https://github.com/AaltoVision/DGC-GNN-release}.}
\end{abstract}
% \thanks{The code and trained models are available at: \url{https://github.com/AaltoVision/DGC-GNN-release}}

\section{Introduction}
% 2D-3D descriptor-based matching
Establishing 2D-3D matches plays a crucial role in various computer vision applications, including visual localization~\cite{irschara2009structure, sattler2012improving, sattler2018benchmarking, sarlin2019coarse, wang2021continual, sarlin2020superglue}, 3D reconstruction~\cite{snavely2006photo, cui2015global, schoenberger2016sfm, lindenberger2021pixel}, and Simultaneous Localization and Mapping (SLAM)~\cite{eade2006scalable, mur2015orb, mur2017orb}. 
Traditional methods for establishing point-to-point matches involve extracting keypoints and descriptors from a query image, then matching the 2D and 3D descriptors using exhaustive search.
To circumvent the computationally expensive matching process, some approaches~\cite{irschara2009structure, sarlin2019coarse} narrow the search space by employing image retrieval methods~\cite{nister2006scalable, Arandjelovic16} first to identify the most similar images in the database, and then perform descriptor-based image matching~\cite{lowe2004distinctive, detone2018superpoint,dusmanu2019d2, sarlin2020superglue, shi2022clustergnn} between the query and retrieved images.
The 2D-3D correspondences are subsequently established by connecting the 2D-2D image matches with the prebuilt 2D-3D correspondences in the database.
Another approach~\cite{sattler2011fast} is to build 2D-to-3D matches by searching through all point descriptors with an efficient vocabulary-based method.
Sattler \emph{et al.}~\cite{sattler2012improving, sattler2016efficient} further explore the combination of both 2D-3D and 3D-2D search as an active correspondence search step for a faster and more efficient matching process.

\begin{figure}[t!]
  \centering
  \includegraphics[width=0.99\linewidth]{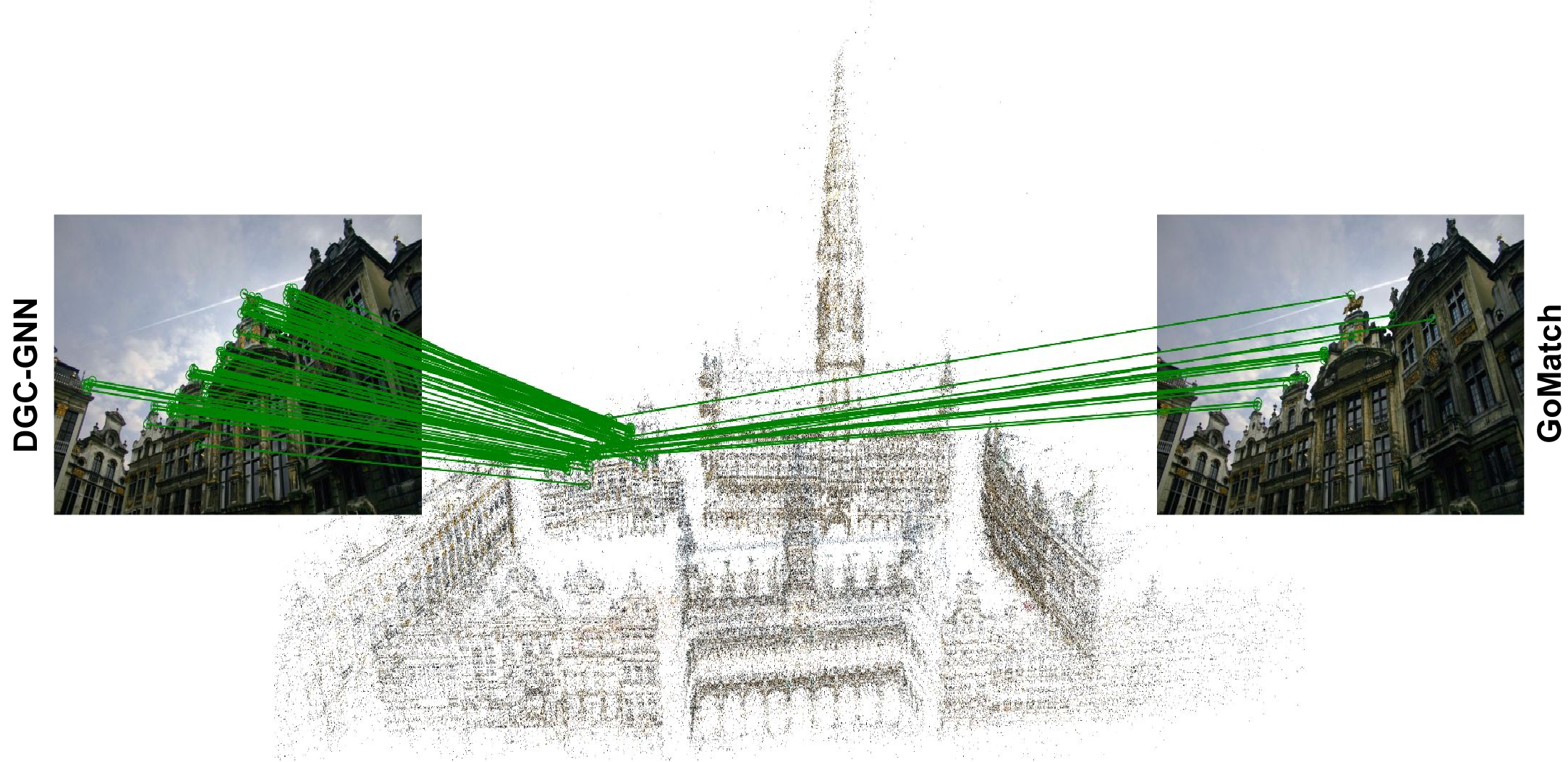}
  \vspace{5pt}
  \caption{\textbf{2D-3D matching} (shown by green lines) with the proposed DGC-GNN and GoMatch~\cite{zhou2022geometry}. 
  In this example, DGC-GNN obtains 78 correct matches with 0.02 meters camera translation and 0.24$^\circ$ rotation errors, while GoMatch finds only 17 inliers with a pose error of 0.37 meters and 4.37$^\circ$. }
  \label{teaser}
\end{figure}
% Other, descriptor-free, approaches
While descriptor-based algorithms achieve state-of-the-art accuracy, they store and maintain high-dimensional visual descriptors for each point in potentially large 3D point clouds. 
The stored model often requires orders of magnitude more storage than the point cloud and images alone~\cite{zhou2022geometry}.
These methods are susceptible to privacy attacks~\cite{dosovitskiy2016inverting, dosovitskiy2016generating, chelani2021privacy, pan2023privacy} and necessitate computationally expensive model maintenance and descriptor update procedures~\cite{zhou2022geometry} when incorporating new descriptors or points into the model.
Several approaches have been proposed to address these limitations. 
Yang \emph{et al.}~\cite{Yang_2022_CVPR} employ learned point selection to sample a subset of the point cloud for scene compression. Other methods~\cite{li2020hierarchical, brachmann2021dsacstar} directly learn a function that maps 2D pixels to 3D coordinates without explicitly storing the 3D scene. Additionally, \cite{ng2022ninjadesc} introduces an adversarial learning framework to develop content-concealing descriptors that prevent privacy leakage.

Recently, researchers~\cite{campbell2020solving, liu2020learning} have begun exploring deep learning techniques for cross-domain direct 2D-3D matching and pose estimation without visual descriptors, showcasing the potential of descriptor-free matching through differentiable geometric optimization.
The recently proposed GoMatch~\cite{zhou2022geometry} represents significant progress in descriptor-free 2D-3D keypoints matching, achieving reasonable matching performance on a variety of real-world datasets~\cite{li2018megadepth, shotton2013scene, kendall2015posenet}.
GoMatch first identifies keypoints in the query image, which, along with the 3D points from the model, are converted to bearing vectors in the camera coordinate system.
The algorithm employs an attention mechanism~\cite{sarlin2020superglue, vaswani2017attention} to establish reliable 2D-3D correspondences effectively.
While GoMatch attains reasonable accuracy, its performance still significantly lags behind its descriptor-based counterparts~\cite{sarlin2020superglue, sarlin2019coarse, sattler2016efficient}.
Additionally, it relies on geometric cues only from the points and their local neighbors, rendering it incapable of distinguishing geometrically similar structures.

These observations lead us to two critical questions: 
(1) Is geometry the only information we can utilize? 
(2) How can we leverage the geometric information derived from the points for matching? 
In practice, humans identify correspondences between objects by considering global structures and local geometric cues. 
For example, when matching an image to a point cloud as in Fig.~\ref{teaser}, we first locate the building based on its unique structure and then identify the local structure of the roof for matching. 
Besides geometric cues, the visual context, such as the color information at each point, also provides constraints for 2D-3D matching. 
Importantly, this color information still preserves privacy, as the RGB data from sparse keypoints is insufficient to reconstruct the scene.

Building upon these observations and the groundwork set by GoMatch, we propose a novel graph-based pipeline, named \textbf{DGC-GNN}, which leverages geometric and color cues in a global-to-local manner for descriptor-free 2D-3D matching. 
DGC-GNN encodes position and RGB information for each point and extracts a global \textit{distance-angular} embedding to guide local point matching. 
Taking inspiration from~\cite{shi2022clustergnn}, we employ a cluster-based transformer to constrain information flow within local clusters. 
We observe, from real-world datasets, that DGC-GNN leads to substantial improvements in the number of correct matches and the accuracy of pose estimation. Notably, it \textit{doubles the accuracy} of GoMatch, thereby reducing the gap between descriptor-based and descriptor-free methods. In summary, our paper makes the following contributions:

\begin{itemize}[leftmargin=0.5cm,noitemsep,topsep=0pt]
\setlength\itemsep{0em}
\item We introduce a visual descriptor-free global-to-local GNN for direct 2D-3D keypoint matching. The network leverages multiple cues and incorporates a progressive clustering module to represent the keypoints. This pipeline enhances the accuracy of sparse 2D-3D matching while requiring low memory, being privacy-preserving, and low cost from 3D model maintenance.
\item We demonstrate that color information for each point is crucial for 2D-3D matching. By incorporating RGB encoding into our network, we observe significant performance improvements.
\item Extensive experiments on real-world datasets show that DGC-GNN outperforms previous methods by a large margin on both matching and visual localization tasks. 
\end{itemize}

\begin{figure*}[t]
  \centering
  \includegraphics[width=0.95\linewidth]{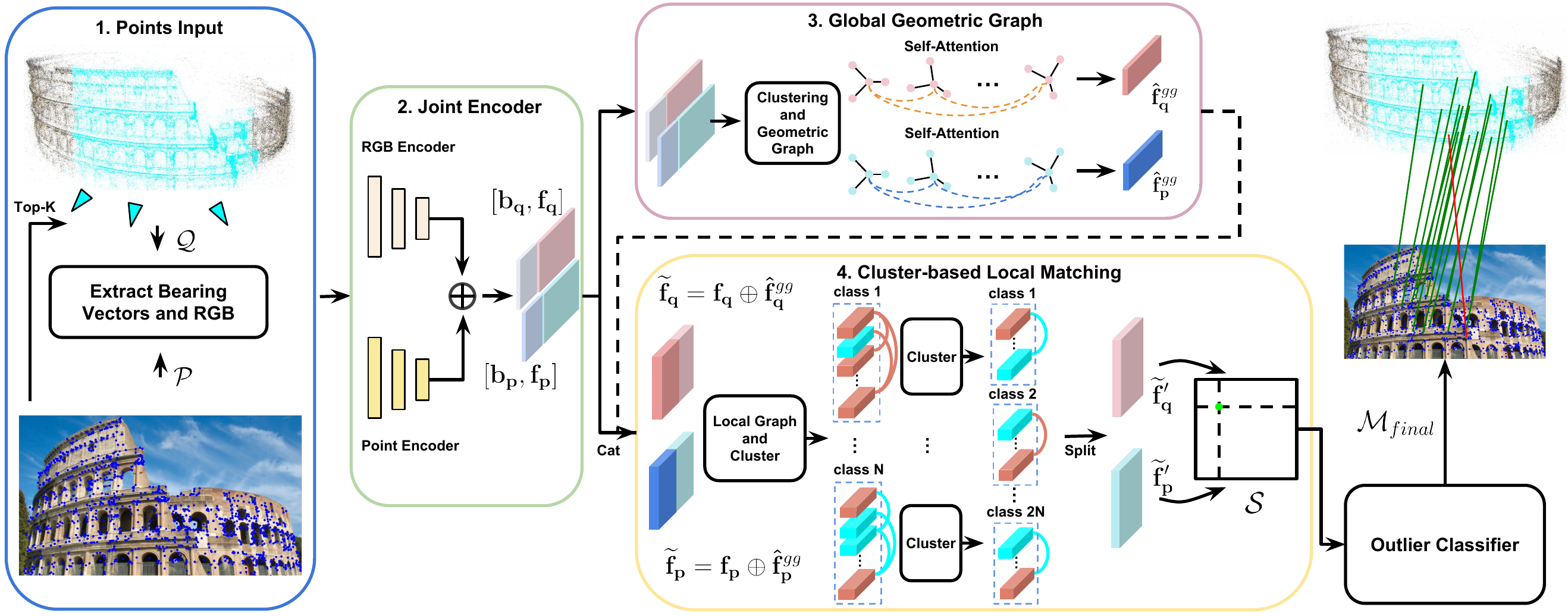}
  \caption{
  \textbf{Pipeline overview.} For keypoints from the 2D image and 3D points from the point cloud, the proposed DGC-GNN (1) considers the bearing vectors and the color at each bearing vector as input.  
  (2) It extracts the point-wise position and color features with two separate encoders and mixes the features as $\mathbf{f_p}$ and $\mathbf{f_q}$. 
  (3) The bearing vectors are clustered into $K$ groups, and geometric graphs are built upon the clusters to extract the global-level geometric embeddings $\mathbf{\hat{f}}_{\mathbf{p}}^{gg}$ and $\mathbf{\hat{f}}_{\mathbf{q}}^{gg}$. 
  (4) We then concatenate $\mathbf{\hat{f}}_{\mathbf{p}}^{gg}$ with $\mathbf{f_p}$ and  $\mathbf{\hat{f}}_{\mathbf{q}}^{gg}$ with $\mathbf{f_p}$, and build a local graph at each point as self-attention. 
  A cluster-based attention module is adopted to enhance the local features by forcing the message to pass only with the most related features. 
  A differentiable layer matches and optimizes the improved features to obtain score matrix $\mathcal{S}$. 
  Finally, an outlier rejection network is applied to prune the matches with low confidence, leading to the final 2D-3D correspondences $\mathcal{M}_{final}$.
  }
  \label{overall}
\end{figure*}

\section{Visual Descriptor-Free 2D-3D Matching}
\subsection{Problem Formulation and Notation}
Given keypoints $ \mathbf{P} = \{\mathbf{p}_n \in \mathbb{R}^2 \; | \; n = 1,...,N\}$ from query image $I$ and database 3D point cloud $\mathbf{Q} = \{\mathbf{q}_m \in \mathbb{R}^3  \; | \; m = 1,...,M\}$, where, optionally, each 3D point is associated with a visual descriptor $\mathbf{d} \in \mathbb{R}^D$. 
The task is to find a set $\mathcal{M}_{\mathbf{p,q}}$ of corresponding keypoints such that 
    \begin{equation}
        \mathcal{M}_{\mathbf{p,q}} = \{(n,m ) \; | \; || \pi(\mathbf{q}_m, \mathbf{R,t,K}) - \mathbf{p}_n ||_2 \leq \epsilon\},
    \end{equation}
where $\pi(\cdot)$ is a mapping that projects a 3D point $\mathbf{q}_m$ from world coordinates to the image plane, represented by a camera rotation $\mathbf{R} \in \mathbb{R}^{3\times3}$, translation $\mathbf{t} \in \mathbb{R}^3$, and intrinsic parameter matrix $\mathbf{K} \in \mathbb{R}^{3\times3}$.
Parameter $\epsilon \in \mathbb{R}$ is the threshold specified in pixels. 
Additionally, we denote the color of point $\mathbf{p}_n$ as $\mathbf{c}_n = [r, g, b]^\text{T} \in [0, 1]^{3}$.

\noindent
\textbf{Bearing Vector.} Similar to~\cite{zhou2022geometry}, we adopt bearing vectors as keypoint representation for both the 2D and 3D points to alleviate their cross-domain nature and represent them in the same space. 
The bearing vector is the direction from the camera center to a 3D point in the camera coordinate system. 
Given an image, a 2D pixel $\mathbf{p}_n$ is uplifted to bearing vector as $[\mathbf{b}_{\mathbf{p},n}, 1]^\text{T} = \mathbf{K^{-1}}[\mathbf{p}_n, 1]^\text{T} ,\mathbf{b}_{\mathbf{p},n} \in \mathbb{R}^2 $, where $\mathbf{K}$ is the intrinsic camera matrix.
Given a 3D point $\mathbf{q}_m$, the corresponding bearing vector is 
\begin{equation}
    [\mathbf{b}_{\mathbf{q},m}, 1]^\text{T} = \frac{\mathbf{Rq}_m + \mathbf{t}}{[\mathbf{Rq}_m + \mathbf{t}]_z},
\end{equation}
where $\mathbf{R}$ is the camera rotation and $\mathbf{t}$ is its translation in the world coordinate system and subscript $z$ denotes the third component of the 3D vector.

\subsection{Network Architecture}
The proposed DGC-GNN applies a hierarchical mechanism to leverage color and geometric cues in a global-to-local fashion. The overall pipeline is illustrated in Fig.\ref{overall}. 
We initially employ two local feature extractors to encode RGB and position information for each point simultaneously (Sec.~\ref{encoding}). Additionally, we cluster the points based on their distances and generate global graphs to obtain the global-level geometric embeddings (Sec.~\ref{coarse}). Next, we concatenate the local point features with their corresponding global features and input them into the cluster-based local matching module to identify the initial matches (Sec.~\ref{attention}). Finally, we incorporate a classification network to filter out matches with low confidence to refine the initial matches (Sec.\ref{pruning}).

\subsubsection{Local Feature Extraction} \label{encoding}
To extract points-wise features from both the 2D keypoint set $\mathbf{P}$ and the 3D point cloud $\mathbf{Q}$, we consider the inputs as bearing vectors equipped with color information: $\mathcal{P} = \{\mathbf{b_p, c_p}\}$ and $\mathcal{Q} = \{\mathbf{b_q, c_q}\}$. Two ResNet-style point encoders~\cite{he2016deep,campbell2020solving}, denoted as $\mathcal{F}_b$ and $\mathcal{F}_c$, are applied to extract position and color embeddings separately. We then obtain the local point features, $\mathbf{f_p}$ and $\mathbf{f_q}$, as follows:
\begin{equation}
    \mathbf{f_p} = \mathcal{F}_b (\mathbf{b_p}) + \mathcal{F}_c (\mathbf{c_p}), \ \  \mathbf{f_q} = \mathcal{F}_b (\mathbf{b_q}) + \mathcal{F}_c (\mathbf{c_q}).
\end{equation}
The resulting point-wise features $\mathbf{f_p}$ and $\mathbf{f_q}$ are vectors with dimensions $\mathbb{R}^{N \times d}$ and $\mathbb{R}^{M \times d}$ respectively, where $N$ and $M$ represent the number of keypoints in $\mathbf{P}$ and $\mathbf{Q}$, and $d$ denotes the dimensionality of the encoded features, e.g., $d = 128$ .
% for SIFT features~\cite{lowe2004distinctive}.

\subsubsection{Global Geometric Guidance}\label{coarse}

\begin{figure*}[t]
  \centering
  \includegraphics[width=0.95\linewidth]{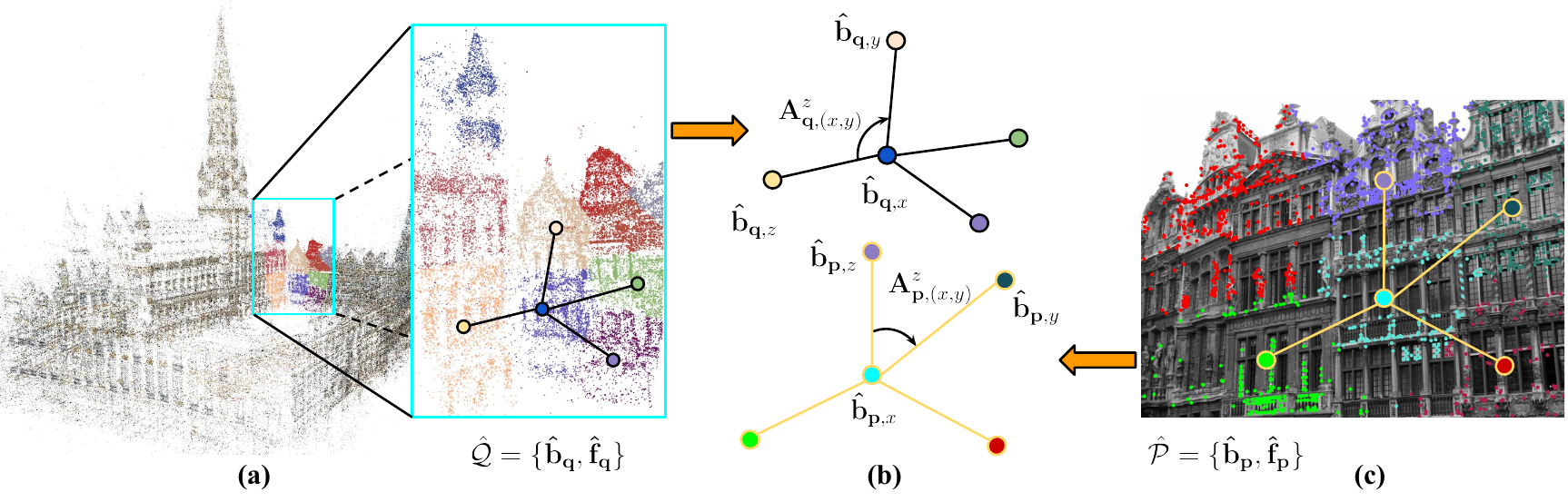}
  \caption{\textbf{Cluster-based geometric encoding}. (a) 
  The clusters obtained from bearing vectors $\mathcal{Q}$ of the 3D point cloud are visualized by color.
  The local graph is created from the neighboring cluster centers. 
  Black 3D points are filtered out from matching. 
  (b) Angular embedding from the global graph to obtain rotation-invariant geometric cues. (c) The clusters obtained from 2D keypoints' bearing vectors $\mathcal{P}$. Similarly, as in 3D, the local graph is created from the neighboring cluster centers. 
  }
  \label{global_graph}

\end{figure*}

Global context guidance has demonstrated its effectiveness in various computer vision tasks~\cite{sun2021loftr, li2020hierarchical, yu2021cofinet, qin2022geometric}. Global context helps to differentiate local descriptors from similar structures or patches, thereby reducing ambiguity. However, most existing methods~\cite{sun2021loftr, qin2022geometric} consider the outputs from different encoding layers as global and local features. This approach is not suitable for our scenario, as our input is sparse points. Downsampling the sparse point cloud results in losing distinctive geometric structures. Hence, we adopt cluster-based geometric encoding to extract global embeddings. 
As shown in Fig.~\ref{global_graph}.(a) and (c), the input bearing vectors, both in the image and in the point cloud, are first clustered into $X$ groups. 
The groups represent distinct clusters, each associated with a cluster center as the global position, denoted by $\mathbf{\hat{b}}_{\mathbf{p},x} \in \mathbb{R}^2, x = 1,...X$. 
The corresponding global embedding is obtained as the average of the point embeddings within a cluster as $\mathbf{\hat{f}}_{\mathbf{p},x} = \frac{1}{P'}\sum_{p'=1}^{P'} \mathbf{f}_{\mathbf{p},p'}$, where  $\mathbf{\hat{f}}_{\mathbf{p},x} \in \mathbb{R}^d$, $P'$ is the number of points in the $x$. The same is conducted on the 3D points to obtain $\mathbf{\hat{b}}_{\mathbf{q}}$ and $\mathbf{\hat{f}}_{\mathbf{q}}$.

\noindent
\textbf{Global Geometric Graph.} To aggregate and extract the geometric relations among the clusters, we propose a novel graph neural network that encodes both distance and angular cues; the basic GNN structure is built upon~\cite{huang2021predator, zhou2022geometry}. 
In the following, we describe the graph construction for the 2D global points set $\mathcal{\hat{P}} = \{\mathbf{\hat{b}_p}, \mathbf{\hat{f}_p}\}$ and the same goes for $\mathcal{\hat{Q}} = \{\mathbf{\hat{b}_q}, \mathbf{\hat{f}_q}\}$.
Each cluster center point $\mathbf{\hat{b}}_{\mathbf{p},x}$ is connected to its $k$-NN neighbours ($k \le K$) in the coordinate space, and $\xi_{\mathbf{p}, (x, y)}$ is the edge between center points $\mathbf{\hat{b}}_{\mathbf{p},x}$ and $\mathbf{\hat{b}}_{\mathbf{p},y}$. We update the feature $\mathbf{\hat{f}}_{\mathbf{p},x}$ using the following equation:
\begin{equation}
    ^{(t+1)}\mathbf{\hat{f}}_{\mathbf{p},x} = \max_{\xi_{\mathbf{p}, (x, y)}} \mathcal{H}_{g1}(^{(t)}\mathbf{\hat{f}}_{\mathbf{p},x} \oplus (^{(t)}\mathbf{\hat{f}}_{\mathbf{p},x} - ^{(t)}\mathbf{\hat{f}}_{\mathbf{p},y}))
    \label{gnn},
\end{equation}
where the $\oplus$ denotes concatenation and $\mathcal{H}_{g1}(*)$ is the linear projection with instance normalization~\cite{ulyanov2016instance} and a LeakyReLU function~\cite{maas2013rectifier}. The $max$ operator applies to the $k$-NN neighbors. The global feature $\mathbf{\hat{f}}_{\mathbf{p},x}$ is updated twice, and calculated as 
\begin{equation}
\mathbf{\hat{f}}_{\mathbf{p},x}^g = \mathcal{H}_{g2}(^{(0)}\mathbf{\hat{f}}_{\mathbf{p},x} \oplus ^{(1)}\mathbf{\hat{f}}_{\mathbf{p},x} \oplus ^{(2)}\mathbf{\hat{f}}_{\mathbf{p},x} ). 
\label{gnn-multi}
\end{equation}
$\mathcal{H}_{g2}$ has a similar structure to $\mathcal{H}_{g1}$, but without shared weights.  Besides the distance embedding, inspired by~\cite{qin2022geometric}, we also adopt the angular embedding to obtain rotation-invariant geometric cues for the global representation. 
To do so, we define the embedding on cluster triplets as shown in Fig.~\ref{global_graph}. (b). 
Given bearing vector $\mathbf{\hat{b}}_{\mathbf{p},x}$ and two of its neighbors $\mathbf{\hat{b}}_{\mathbf{p},y}$ and $\mathbf{\hat{b}}_{\mathbf{p},z}$,
the angular embedding of <$\mathbf{\hat{b}}_{\mathbf{p},x}$, $\mathbf{\hat{b}}_{\mathbf{p},y}$> \wrt $\mathbf{\hat{b}}_{\mathbf{p},z}$ is defined as follows:
 \begin{equation}
     \mathbf{A}_{x,y}^z = \sine (\angle(\mathbf{\hat{b}}_{\mathbf{p},z} - \mathbf{\hat{b}}_{\mathbf{p},x}, \ \ \ \mathbf{\hat{b}}_{\mathbf{p},y} - \mathbf{\hat{b}}_{\mathbf{p},x}) / \sigma_a ),
 \end{equation}
where $\sine(\cdot)$ is a sinusoidal function and $\sigma_a$ is a controller constant, 
all $k$ neighbours are considered to obtain the angular embedding $\mathbf{A}_{\mathbf{p}}$. 
We update the \textit{global geometric embedding} $\mathbf{\hat{f}}_{\mathbf{p}}^{gg}$ as an angular-aware attention mechanism:
\begin{equation}
\label{angular-atten}
    \mathbf{\hat{f}}_{\mathbf{p}}^{gg} = \norm (\mathbf{\hat{f}}_{\mathbf{p}}^g + \Att(\mathbf{\hat{f}}_{\mathbf{p}}^g, \mathbf{A}_{\mathbf{p}}));  \ \ \ \  \mathbf{\hat{f}}_{\mathbf{p}}^{gg} \in \mathbb{R}^{K\times d},
\end{equation}
where
\vspace{-5pt}
\begin{equation*}
\begin{split}
        \Att(\mathbf{\hat{f}}_{\mathbf{p}}^g, \mathbf{A}_{\mathbf{p}}) = \\(\mathbf{\hat{f}}_{\mathbf{p}}^g \mathbf{W^V}) . \frac{(\mathbf{A}_{\mathbf{p}}\mathbf{W^A})(\mathbf{\hat{f}}_{\mathbf{p}}^g \mathbf{W^Q})^\text{T} + (\mathbf{\hat{f}}_{\mathbf{p}}^g \mathbf{W^Q})(\mathbf{\hat{f}}_{\mathbf{p}}^g \mathbf{W^K})^\text{T}}{\sqrt{dim}}.
\end{split}
\end{equation*}
$\mathbf{W^A}, \mathbf{W^Q}, \mathbf{W^K}, \mathbf{W^V} \in \mathbb{R}^{d \times d}$ are the projection matrices of each item and LayerNorm~\cite{ba2016layer} is applied to Eq.~\ref{angular-atten}. 
Each local point feature is associated with its corresponding global embedding by $\mathbf{\widetilde{f}}_{\mathbf{p}} = \mathbf{f}_{\mathbf{p}} \oplus \mathbf{\hat{f}}_{\mathbf{p}}^{gg}, \mathbf{\widetilde{f}}_{\mathbf{p}} \in \mathbb{R}^{N\times2d}$ to obtain $\mathcal{\widetilde{P}} = \{ \mathbf{b}_{\mathbf{p}}, \mathbf{\widetilde{f}}_{\mathbf{p}}\}$.
The same procedure obtains local and global embedding $\mathcal{\widetilde{Q}} = \{ \mathbf{b}_{\mathbf{q}}, \mathbf{\widetilde{f}}_{\mathbf{q}}\}$ for the point cloud $\mathcal{\hat{Q}}$.

\subsubsection{Cluster-based Local Matching}\label{attention}
After extracting the global geometric embedding, we implement a \textit{cluster-based matching module} to obtain the initial intra-domain 2D-3D matches. This cluster-based GNN ~\cite{shi2022clustergnn} has been shown to be more computationally efficient than its complete-graph counterpart~\cite{sarlin2020superglue}. 
The network considers the local point features from both $\mathcal{\widetilde{P}}$ and $\mathcal{\widetilde{Q}}$ a complete set, then clusters the feature with strong correlations into the same group and restricts the message passing within each group. 
In addition to its low computational complexity, we found that cluster GNN can effectively utilize global-to-local geometric cues, as the clustering operation inherits the property of global graph clustering and forces it to distinguish ambiguous local features even with similar global embedding.

\begin{table*}[t!]
\begin{center}
\renewcommand\arraystretch{1.5}
\resizebox{.99\textwidth}{!}
{
\huge
\begin{tabular}{llccccccccc}
\hline \specialrule{2.5pt}{0.5pt}{0.5pt}
\multicolumn{2}{l}{\multirow{3}{*}{Methods}} &
  \multicolumn{4}{c}{ScanNet~\cite{dai2017scannet}} &
  \textbf{} &
  \multicolumn{4}{c}{MegaDepth~\cite{li2018megadepth}} \\ \cline{3-6} \cline{8-11} 
\multicolumn{2}{l}{} &
  Reproj. AUC (\%) &
  Rotation ($^\circ$)  &
  Translation (m) &
  \multirow{2}{*}{P (\%)  ($\uparrow$)}  &
  \multirow{2}{*}{\textbf{}} &
  Reproj. AUC (\%) &
  Rotation ($^\circ$) &
  Translation 
  & \multirow{2}{*}{P (\%) ($\uparrow$)} \\
\multicolumn{2}{l}{} &
  @1 / 5 / 10px ($\uparrow$) &
  \multicolumn{2}{c}{Quantile @25 / 50 / 75\% $(\downarrow)$} &
   & &
  @1 / 5 / 10px ($\uparrow$) &
  \multicolumn{2}{c}{Quantile @25 / 50 / 75\% $(\downarrow)$} \\ \hline \specialrule{1.5pt}{0.5pt}{0.5pt}
\multirow{4}{*}{k=1} &
Oracle & 29.13 / 39.83 / 41.34 & \phantom{11}0.10 / \phantom{11}0.19 / \phantom{11}0.40 & 0.01 / 0.01 / 0.03&-&& 34.59 / 85.02 / 92.02 & \phantom{1}0.04 / \phantom{1}0.06 / \phantom{1}0.12& 0.00 / 0.01 / 0.01 &- \\
&
  BPnPNet~\cite{campbell2020solving} &
  \phantom{1}0.00 / \phantom{1}0.00 / \phantom{1}0.02 &
  \phantom{1}99.17 / 128.90 /  154.68 &
  4.35 / 6.82 / 9.86 & \phantom{1}3.60 &
  \multicolumn{1}{l}{} &
  \phantom{1}0.22 / \phantom{1}0.63 / \phantom{1}0.89 &
  16.13 / 32.01 / 61.58 &
  1.67 / 3.17 / 5.44  & \phantom{1}2.95\\
 &
  GoMatch~\cite{zhou2022geometry} &
  \phantom{1}1.18 / 11.23 / 18.01 &
  \phantom{11}2.69 / \phantom{1}12.78 / \phantom{1}36.50 &
  0.19 / 0.91 / 2.63 & 13.18 &
  \multicolumn{1}{l}{} &
  \phantom{1}5.67 / 22.43 / 28.01 &
  \phantom{1}0.60 / 10.08 / 34.63 & 
  0.06 / 1.06 / 3.73 & 14.94\\ \cline{3-11} 
 &
  \textbf{DGC-GNN} &
  \textbf{\phantom{1}2.73 / 21.88 / 32.23} &
  \textbf{\phantom{11}0.94 / \phantom{11}3.17 / \phantom{1}20.14} &
  \textbf{0.06} / \textbf{0.23} / \textbf{1.40} & \textbf{14.86} &
  \multicolumn{1}{l}{} &
  \textbf{10.20 / 37.64 / 44.04} &
  \phantom{1}\textbf{0.15} / \phantom{1}\textbf{1.53} / \textbf{27.93} &
  \textbf{0.01 / 0.15 / 3.00} & \textbf{19.00} \\\cline{1-11} 
\multirow{3}{*}{k=10} &
  BPnPNet~\cite{campbell2020solving} &
  \phantom{1}0.00 / \phantom{1}0.00 / \phantom{1}0.03 &
  104.68 / 135.94 / 160.54 &
  4.67 / 7.30 / 10.92 & \phantom{1}0.84 &
   &
\phantom{1}0.36 /   \phantom{1}0.72 /   \phantom{1}0.97 &
  16.63 / 34.69 / 67.77 &
  1.64 / 3.30 / 5.97  & \phantom{1}0.74\\
 &
  GoMatch~\cite{zhou2022geometry} &
  \phantom{1}0.91 / 18.98 / 31.12 &
  \phantom{11}1.18 /  \phantom{11}4.94 / \phantom{1}28.97 &
  \phantom{1}0.08 / 0.35 / 2.08 & \phantom{1}4.25 &
   &
  \phantom{1}8.90 / 35.67 / 44.99 &
  \phantom{1}0.18 /   \phantom{1}1.29 / 16.65 &
  0.02 / 0.12 / 1.92  & \phantom{1}8.76 \\ \cline{3-11} 
 &
  \textbf{DGC-GNN} &
  \phantom{1}\textbf{1.76} / \textbf{31.74} / \textbf{48.11} &
  \phantom{11}\textbf{0.67}  / \phantom{11}\textbf{1.49} /  \phantom{11}\textbf{7.62} &
  \phantom{1}\textbf{0.04} / \textbf{0.11} / \textbf{0.53} & \phantom{1}\textbf{6.42} &
  \multicolumn{1}{l}{} &
  \textbf{15.30 /  51.70 /  60.01} &
 \phantom{1}\textbf{0.07} / \phantom{1}\textbf{0.26} / \phantom{1}\textbf{5.41} &
  \textbf{0.01 / 0.02 / 0.57} & \textbf{13.36} \\ \hline \specialrule{2pt}{0.5pt}{0.5pt}
\end{tabular}}
\caption{\textbf{2D-3D Matching.} 
We present AUC scores for reprojection errors threshold at 1, 5, and 10 pixels; rotation and translation error quantiles at 25, 50, and 75\%; and matching precision. 
Parameter $k$ is the number of images retrieved from the database to narrow down the search space. 
The best results are bold.
DGC-GNN nearly doubles the AUC scores of GoMatch and reduces the pose errors to their ${\approx}33\%$. 
} 
\vspace{-10pt}
\label{tab:matching}
\end{center}
\end{table*}
\vspace{-5pt}
\vspace{5pt}
\noindent
\textbf{Initialization.} 
As an initialization procedure for the cluster attention module, we run the general self and cross-attention modules proposed in GoMatch~\cite{zhou2022geometry}.
For each local point $\mathbf{b}_{\mathbf{p}, n}$, we construct a local graph according to its $k'$ nearest neighbours in the Euclidean space and update the associated feature $\mathbf{\widetilde{f}}_{\mathbf{p}, n} \in \mathbb{R}^{2d}$ by Eq.~\ref{gnn-multi}. 
Note that we ignore the angular embedding at this stage due to the unaffordable memory requirements with space complexity $\mathcal{O}(Nk'^2)$, where $N$ is the number of local points. We then use linear attention~\cite{katharopoulos2020transformers, sun2021loftr} as a cross-attention mechanism, which allows each point in one modality to interact with all points from another modality. This not only facilitates inter-modality in the feature matching but also reduces the computational complexity from $\mathcal{O}(N^2)$ to $\mathcal{O}(N)$.

\noindent
\textbf{Cluster-based Attention.} After the graph initialization, the features $\mathbf{\widetilde{f}}_{\mathbf{p}}$ and $\mathbf{\widetilde{f}}_{\mathbf{q}}$ coming from the image and the point cloud respectively, are concatenated and processed in a two-level hierarchical clustering attention module. 
The hierarchical structure is effective in suppressing erroneous groupings. At the first level, we cluster the feature vectors into $I$ coarse groups. 
In the second level, each coarse group is divided into several small groups. 
The local point information exchange is conducted at each level and only within each group to obtain more representative features. 
After the sparse clustering, each feature vector is transformed back to its original position and then split again into $\mathbf{\widetilde{f}'}_{\mathbf{p}}$ and $\mathbf{\widetilde{f}'}_{\mathbf{q}}$ to obtain the keypoints both in the 2D and 3D spaces.

\noindent
\textbf{Optimal Transport.} We calculate the cost matrix $\mathcal{M} \in \mathbb{R}^{N \times M}$ between the two transformed feature sets using the $L_2$ distance between pairs of features. Thus, $\mathcal{M}(n,m) = ||\mathbf{\widetilde{f}'}_{\mathbf{p}, n}-  \mathbf{\widetilde{f}'}_{\mathbf{q}, m}||_2$. Following~\cite{sarlin2020superglue}, the cost matrix $\mathcal{M}$ is extended to $\mathcal{\bar{M}}$ by adding an additional row and column as dustbins for unmatched points. 
We then iteratively optimize $\mathcal{\bar{M}}$ running the Sinkhorn algorithm~\cite{sinkhorn1967concerning, cuturi2013sinkhorn} in a declarative layer to obtain the score matrix $\mathcal{\bar{S}}$.
Finally, $\mathcal{\bar{S}}$ is converted to $\mathcal{S} \in \mathbb{R}^{N \times M}$ by dropping the dustbins. 
The initial 2D-3D match candidates are acquired by mutual top-1 search, thus
\begin{equation}
    \mathcal{M}_{init} = \{(\widetilde{n}, \widetilde{m}) \ | \ \forall (\widetilde{n}, \widetilde{m}) \in \text{MNN}(\mathcal{S}) \},
\end{equation}
where $\text{MNN}$ is the mutual nearest neighbors operator. %, keeping only those matches that are mutually the top-1 matches of each other.
Set $\mathcal{M}_{init}$ provides initial 2D-3D matches that we further filter in Sec.~\ref{pruning} to keep the accurate correspondences only. 

\subsection{Outlier Rejection} \label{pruning}

After obtaining the initial matches, outlier pruning runs to remove the incorrect ones. We apply the same outlier rejection network as in GoMatch~\cite{zhou2022geometry}, whose input is the concatenated 2D and 3D keypoint features $\mathbf{\widetilde{f}'}_{\widetilde{n}, \widetilde{m}} = \mathbf{\widetilde{f}'}_{\mathbf{p}, \widetilde{n}} \oplus \mathbf{\widetilde{f}'}_{\mathbf{q}, \widetilde{m}}$ and outputs the matching confidence of each matched pair. 
The final predicted matches are obtained as follows: 
\begin{equation}
        \mathcal{M}_{final} = \{(\widetilde{n}', \widetilde{m}') \ |  \forall \  \text{cls}(\mathbf{\widetilde{f}'}_{\widetilde{n}, \widetilde{m}} \; | \; (\widetilde{n}, \widetilde{m}) \in \mathcal{M}_{init}) \geq \theta  \},
\end{equation}
where $\theta$ is the matching confidence threshold. 

\subsection{Training Loss}
We use the same training loss as GoMatch. 
The loss function $\mathcal{L}$ consists of two terms, the matching loss $\mathcal{L}_{ot}$ and the classification loss $\mathcal{L}_{or}$. The ground truth match set $\mathcal{M}_{gt}$ is estimated by reprojecting the 3D points to the 2D image plane and calculating the pixel distance. We also include point sets $\mathcal{I}$ and $\mathcal{J}$ for the unmatched points in $\mathcal{P}$ and $\mathcal{Q}$, respectively. The matching loss $\mathcal{L}_{ot}$ minimizes
the negative log-likelihood of the matching score $\mathcal{\bar{S}}$.
\begin{equation}
\begin{split}
    \mathcal{L}_{ot} = - \frac{1}{|\mathcal{M}_{gt}| + |\mathcal{I}| + |\mathcal{J}|}(\sum\limits_{(n,m)\in \mathcal{M}_{gt}} \log \mathcal{\bar{S}}_{n,m} + \\
    \sum\limits_{i \in \mathcal{I}} \log \mathcal{\bar{S}}_{i,m+1} + \sum\limits_{j \in \mathcal{J}} \log \mathcal{\bar{S}}_{N+1,j}).
\end{split}
\end{equation}
The classification loss is defined as 
\begin{equation}
    \mathcal{L}_{or} =  -\frac{1}{|\mathcal{M}_{init}|}\sum_{i=1}^{|\mathcal{M}_{init}|}w_i(y_i\log(p_i) + (1-y_i)\log(1-p_i)),
\end{equation}
where $w_i$ is the weight balancing the positive and negative samples, $y_i$ is the ground truth matching label for the $i$-th correspondences, $p_i$ is the predicted probability of a true match for the $i$-th correspondences. The total loss is the sum of the two terms as $\mathcal{L} = \mathcal{L}_{ot} + \mathcal{L}_{or}$.

\section{Experiments}
\label{results}
% In this section, we first describe the experimental setup, employed datasets, and evaluation protocols. Next, we provide detailed comparisons with the baselines on both matching and visual localization tasks. Finally, we conduct ablation studies on each designed component.

\noindent
\textbf{Training.}
We train the indoor model of DGC-GNN on the ScanNet~\cite{dai2017scannet} dataset and the outdoor model on the MegaDepth~\cite{li2018megadepth} dataset. 
We extract up to 1024 keypoints for each training image by the SIFT detector~\cite{lowe2004distinctive}. 
Similarly as in GoMatch, we first select a subset of the point cloud by applying image retrieval approaches~\cite{Arandjelovic16, torii201524} to obtain potential images observing the same part of the scene as the input one.
We randomly sample the retrieval pairs with a visual overlap of more than 35\% on MegaDepth and 65\% on ScanNet to ensure enough matches on each pair.  
For the global geometric embedding, we cluster the 2D/3D bearing vectors into $X = 10$ groups, and each cluster center is connected to its $k=4$ nearest neighbors to build the global graph. 
For the local point graph, we connect each point with its 10 nearest neighbors and the cluster-based attentions are performed twice to force the intra-cluster information exchange. 

We use Adam optimizer with a learning rate of 1e-3. We train DGC-GNN with one 32GB Telsa V100 GPU. 
The convergence of the model typically requires 50 epochs. 
% for both indoor and outdoor scenes.
\vspace{1mm}

\noindent
\textbf{Datasets.} 
We use ScanNet and MegaDepth for training and 2D-3D matching task evaluation. 
As a downstream application, we perform visual localization on the 7Scenes~\cite{shotton2013scene} and Cambridge Landmarks~\cite{kendall2015posenet} datasets. 
MegaDepth is a popular outdoor dataset with 196 scenes captured around the world. 
The sparse 3D reconstructions are provided by the COLMAP~\cite{schoenberger2016sfm} structure-from-motion software.  
Following~\cite{zhou2022geometry}, we train our outdoor model on 99 scenes and evaluate it with 53 scenes. 
ScanNet is a large-scale RGB-D indoor dataset comprising 1613 scans with over 2.5 million images. 
We randomly selected 105 scenes for the training and 30 for the evaluation. Cambridge Landmarks is a middle-scale outdoor dataset consisting of 6 individual scenes. 
A structure-from-motion algorithm provides the ground truth camera poses. 
We follow~\cite{kendall2015posenet, zhou2022geometry} to evaluate our method on four scenes. 
7Scenes is a small indoor dataset with RGB-D images and camera poses provided by the depth SLAM system. 
We evaluate on the standard test sequences. 
\vspace{1mm}

\noindent
\textbf{Evaluation Protocol.}
For matching on ScanNet and MegaDepth, we follow~\cite{zhou2022geometry} and report the AUC score calculated from the reprojection errors.
To calculate the errors for the 2D-3D matches in $\mathcal{M}_{final}$, we project the 3D points to the image plane using the ground truth and estimated camera poses. 
Then, we calculate the $L_2$ distance of the ground truth and estimated reprojected 2D points. 
We use multiple thresholds, 1, 5, and 10 pixels, to evaluate the AUC scores.  
The camera translation and rotation error quantiles at 25\%, 50\%, and 75\% are also reported. 
Moreover, we evaluate the matching quality by calculating the matching precision P, which is the ratio of inlier matches after PnP-RANSAC to the number of final matches $\mathcal{M}_{final}$. 
For visual localization tasks, we report the median translation (in meters) and rotation (in degrees) camera pose errors.

\subsection{2D-3D Matching}

We compare with the two descriptor-free matchers GoMatch~\cite{zhou2022geometry} and BPnPnet~\cite{campbell2020solving}. 
At inference, we use the 3D points from the top-$k$ retrieved database images to match with the keypoints from query images. 
Following~\cite{zhou2022geometry}, we report the upper bound of the AUC score using the ground truth matches. We refer to these values as Oracle. 
We select the GT matches by thresholding the reprojection error based on normalized image coordinates, using a threshold of $0.001$, to bypass the influence of camera intrinsics during GT selection. 
This is in contrast to what is done in \cite{zhou2022geometry}.
Results on GT selected by a pixel threshold are in the supp.\ material. 
We use the official code with the default setting to generate the evaluation dataset on MegaDepth~\cite{li2018megadepth} and rerun GoMatch and BPnPNet with the released models.
Note that we also tested GoMatch after retraining it on MegaDepth and achieved similar results as with the released model. 
\vspace{1mm}

\begin{table*}[th]
\begin{center}
\renewcommand\arraystretch{1.3}
\setlength{\tabcolsep}{3pt}
\resizebox{0.98\textwidth}{!}
{
\small
\begin{tabular}{lccccccc}
\hline
\specialrule{0.5pt}{0.5pt}{0.5pt}
\multirow{2}{*}{Methods} &
  G. Emb. &
  C. Att. &
  Color &
  Ang. &
  % \multirow{2}{*}{Cluster} &
  Reproj. AUC (\%) &
  Rotation ($^\circ$) &
  Translation \\
                                        & Sec.~\ref{coarse} & Sec.~\ref{attention} & Sec.~\ref{encoding} &  Sec.~\ref{coarse}             & @1 / 5 / 10px  ($\uparrow$)           & \multicolumn{2}{c}{Quantile@25 / 50 / 75\% ($\downarrow$)}        \\ \hline 
                                        \specialrule{0.5pt}{0.5pt}{0.5pt}
GoMatch~\cite{zhou2022geometry}                   &  &  &  &    & \phantom{1}8.90 / 35.67 / 44.99 & 0.18 / 1.29 / 16.65 & 0.02 / 0.12 / 1.92 \\ \hline

\multirow{5}{*}{Variants}  & \Checkmark &  &  &       & 10.86 / 41.18 / 50.51           & 0.13 / 0.76 / 13.47           & 0.01 / 0.07 / 1.62      \\
                                        & \Checkmark & \Checkmark &  &     &  11.64 / 44.46 / 53.99           &  0.11 / 0.55 / \phantom{1}9.49         &  0.01 / 0.05 / 1.05      \\
                                        &  & \Checkmark & \Checkmark &     &  
                                        
                                        13.20  / 46.33 / 54.34         &    0.09 / 0.41 / \phantom{1}9.98       & 0.01 / 0.03 / 1.19\\        
                                        & \Checkmark & \Checkmark & \Checkmark &      &  14.19 / 48.34 / 56.54           &  0.08 / 0.34 / \phantom{1}9.23         &  0.01 / 0.03 / 1.03      \\
\hline
\textbf{DGC-GNN}                  & \Checkmark & \Checkmark & \Checkmark & \Checkmark & \textbf{15.30 /  51.70 /  60.01} & \textbf{0.07 /  0.26 /  \phantom{1}5.41} & \textbf{0.01 / 0.02 / 0.57} 

\\ \hline
\specialrule{0.5pt}{0.5pt}{0.5pt}
\end{tabular}
}
\caption{\textbf{Ablation Study.}
AUC scores thresholded at 1, 5, and 10 pixels; rotation and translation error quantiles at 25, 50, 75\% with the proposed components added one by one to the GoMatch pipeline on the MegaDepth dataset. 
} 
\vspace{-10pt}
\label{tab:abla}
\end{center}
\end{table*}
\noindent
\textbf{Matching Results.}
The results with $k=1$ and $k=10$ are presented in Table~\ref{tab:matching}. 
Parameter $k$ is the number of retrieved image pairs that are used for evaluation.
The proposed method outperforms GoMatch and BPnPNet by a significant margin on both scenes. 
Specifically, DGC-GNN achieves 10.2 / 37.64 / 44.04\% reprojection AUC compared to GoMatch with 5.67 / 22.43 / 28.01\% on MegaDepth with $k = 1$. 
DGC-GNN halves the rotation and translation errors of GoMatch on all thresholds and it obtains better matching quality. 
Notably, the performance of DGC-GNN with $k=1$ surpasses that of GoMatch with $k=10$, indicating the effectiveness of our method even with a single view.

\begin{figure}[t]
  \centering
  \includegraphics[width=0.99\linewidth]{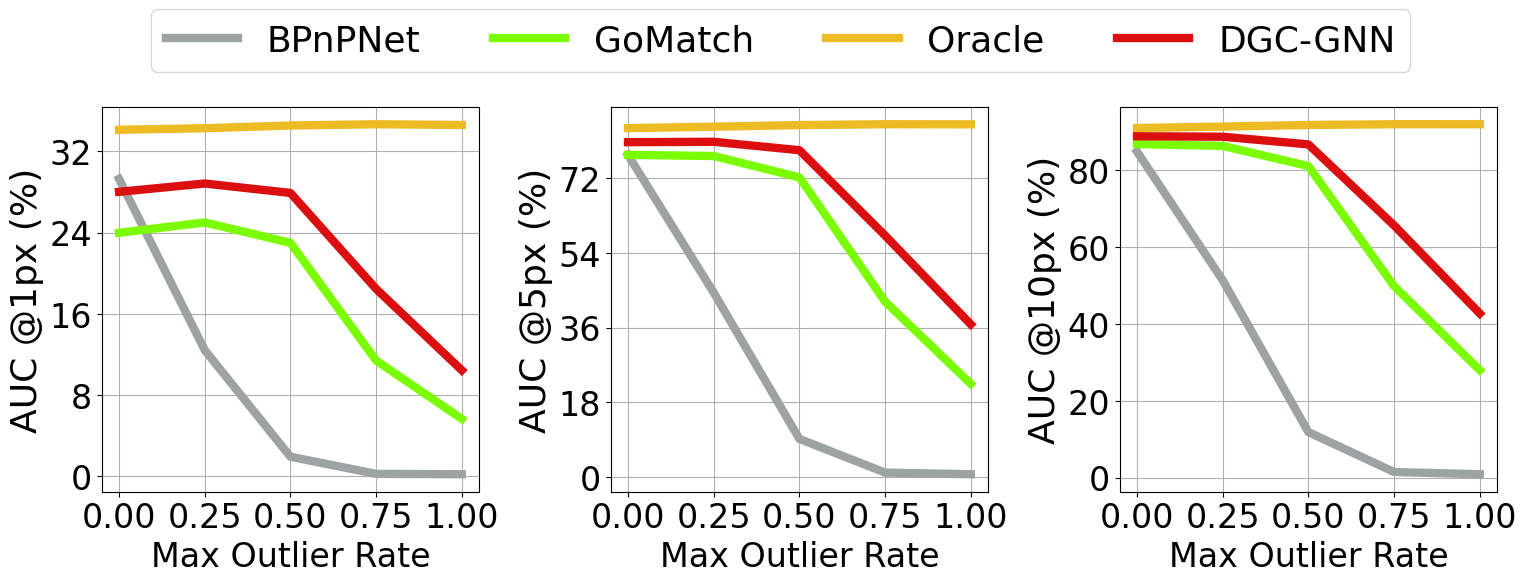}
  \caption{\textbf{Outlier Sensitivity}.
  The AUC scores of BPnPNet~\cite{campbell2020solving}, GoMatch~\cite{zhou2022geometry}, and the proposed DGC-GNN thresholded at 1, 5, and 10 pixels are plotted as a function of the outlier ratio. 
  Oracle represents the AUC upper bound using ground truth matches. 
  }
  \label{outliers}
\end{figure}

\noindent
\textbf{Sensitivity to Outliers.} 
To evaluate the sensitivity to keypoint outliers, we follow the procedure in GoMatch~\cite{zhou2022geometry}. 
The outliers are controlled by the outlier ratio, ranging from 0 to 1, calculated as the number of unmatched keypoints divided by the maximum of the numbers of 2D and 3D points.
If the outlier ratio is $0$, all the input 2D and 3D points are selected from the ground truth matches, and no outliers are included in the matching process. 
When it is $1$, we directly use the keypoints from the query image and 3D points from the top-$k$ retrieved images without any filtering or outlier removal. 
The results are shown in Fig.~\ref{outliers}. 
Even in the presence of outliers, DGC-GNN outperforms other methods by a large margin. 
This indicates that our method is more robust to outliers and can handle challenging matching scenarios more effectively than the state-of-the-art. 

\noindent
\textbf{Ablation Study.} We investigate the effectiveness of different components of DGC-GNN on the 2D-3D matching quality on the MegaDepth dataset~\cite{li2018megadepth} with $k=10$. 
The results are reported in Table~\ref{tab:abla}. 
We provide results with $k=1$ in the supp.\ material. 
We conduct the ablations by gradually adding the components: global geometric embedding (G. Emb), cluster attention (C. Att.), Color, and Angular embedding (Ang.) to the original GoMatch pipeline. 
Incorporating color information into the matching process significantly impacts the performance, resulting in improvements of 2.55 / 3.88 / 2.55\% (AUC@1 / 5 / 10px). 
This demonstrates the importance of color cues for accurate and robust matching.  The global-to-local geometric (G. Emb.) and Angular relation embedding (Ang.) substantially improve the matching performance
by 1.90 / 5.51 / 5.52\% and 1.11 / 3.36 / 3.47\%, respectively. 
It highlights the effectiveness of incorporating global geometric context and local geometric details. 
The cluster attention mechanism also plays a vital role, improving performance by 0.78 / 3.28 / 3.48\%. 
The best results are obtained when all components are added to the pipeline. 

\begin{table*}[t!]
\begin{center}
\renewcommand\arraystretch{1.6}
\resizebox{0.99\textwidth}{!}
{
\Huge

\begin{tabular}{llccccccc|ccccccccc}
\hline \specialrule{2.5pt}{0.5pt}{0.5pt}
\multicolumn{2}{l}{\multirow{2}{*}{{Methods}}} &
  \multicolumn{1}{c}{\multirow{2}{*}{{\begin{tabular}[c]{@{}c@{}}No Desc. \\ Maint.\end{tabular}}}} &
  \multicolumn{1}{c}{\multirow{2}{*}{{Privacy}}} &
  \multicolumn{4}{c}{{Cambridge-Landmarks~\cite{kendall2015posenet} (cm, $^\circ$)}} &
  \multirow{2}{*}{{MB used}} &
  {} &
  \multicolumn{7}{c}{{7Scenes~\cite{shotton2013scene} (cm, $^\circ$)}} &
  \multirow{2}{*}{{MB used}} \\ \cline{5-8} \cline{11-17}
\multicolumn{2}{l}{} &
  \multicolumn{1}{c}{} &
  \multicolumn{1}{c}{} &
  {King’s} &
  {Hospital} &
  {Shop} &
  {St. Mary’s} &
   &
  {} &
  {Chess} &
  {Fire} &
  {Heads} &
  {Office} &
  {Pumpkin} &
  {Kitchen} &
  {Stairs} &
   \\ \hline \specialrule{1.5pt}{0.5pt}{0.5pt}
 \multirow{3}{*}{\rotatebox{90}{{E2E}}} &
  {MS-Trans.~\cite{shavit2021learning}} &  \Checkmark
   &   \Checkmark
   &
  83 / 1.47 &
  181 / 2.39 &
  86 / 3.07 &
  162 / 3.99 &
  \phantom{11}71 &
   &
  11 / 4.66 &
  24 / 9.60 &
  14 / 12.19 &
  17 / 5.66 &
  18 / 4.44 &
  17 / 5.94 &
  26 / 8.45 &
  \phantom{111}71 \\
 &
  {DSAC*~\cite{brachmann2021dsacstar}} &  \Checkmark
   &   \Checkmark
   &
  \textbf{15 / 0.30} &
  \phantom{1}21 / 0.40 &
  \textbf{\phantom{1}5 / 0.30} &
  \phantom{1}13 / 0.40 &
  \phantom{1}112 &
   &
  \phantom{1}\textbf{2} / 1.10 &
  \phantom{1}\textbf{2} / 1.24 &
  \phantom{1}\textbf{1} / 1.82 &
  \phantom{1}\textbf{3} / 1.15 &
  \phantom{1}\textbf{4} / 1.34 &
  \phantom{1}\textbf{4} / 1.68 &
  \phantom{1}\textbf{3} / 1.16 &
  \phantom{11}196 \\
 &
  {HSCNet~\cite{li2020hierarchical}} & \Checkmark
   &   \Checkmark
   &
  18 / \textbf{0.30} &
  \phantom{1}\textbf{19 / 0.30} &
  \phantom{1}6 / \textbf{0.30} &
  \phantom{11}\textbf{9 / 0.30} &
  \phantom{1}592 &
   &
  \phantom{1}\textbf{2 / 0.70} &
  \phantom{1}\textbf{2 / 0.90} &
  \phantom{1}\textbf{1 / 0.90} &
  \phantom{1}\textbf{3 / 0.80} &
  \phantom{1}\textbf{4 / 1.00} &
  \phantom{1}\textbf{4 / 1.20} &
  \phantom{1}\textbf{3 / 0.80} &
  \phantom{1}1036 \\
 \hline
 \multirow{3}{*}{\rotatebox{90}{{DB}}} &
  {HybridSC~\cite{camposeco2019hybrid}} &  \XSolidBrush
   & --
   &
  81 / 0.59 &
  \phantom{1}75 / 1.01 &
  19 / 0.54 &
  \phantom{1}50 / 0.49 &
  \phantom{111}3 &
  \multicolumn{1}{c}{} &
  \multicolumn{1}{c}{-} &
  \multicolumn{1}{c}{-} &
  \multicolumn{1}{c}{-} &
  \multicolumn{1}{c}{-} &
  \multicolumn{1}{c}{-} &
  \multicolumn{1}{c}{-} &
  \multicolumn{1}{c}{-} &
  - \\
 &
  {AS~\cite{sattler2016efficient}}&  \XSolidBrush
  &  \XSolidBrush
   &
  13 / 0.22 &
  \phantom{1}20 / 0.36 &
  \phantom{1}\textbf{4} / 0.21 &
  \phantom{11}8 / 0.25 &
  \phantom{1}813 &
   &
  \phantom{1}3 / 0.87 &
  \phantom{1}\textbf{2} / 1.01 &
  \phantom{1}\textbf{1} / 0.82 &
  \phantom{1}4 / 1.15 &
  \phantom{1}7 / 1.69 &
  \phantom{1}5 / 1.72 &
  \textbf{4} / \textbf{1.01} &
  - \\
 &
  {SP~\cite{detone2018superpoint}+SG~\cite{sarlin2020superglue}} &  \XSolidBrush
    &  \XSolidBrush
   &
  \textbf{12 / 0.20} &
  \phantom{1}\textbf{15 / 0.30} &
  \phantom{1}\textbf{4 / 0.20} &
  \phantom{11}\textbf{7 / 0.21} &
  3215 &
   &
  \phantom{1}\textbf{2 / 0.85} &
  \phantom{1}\textbf{2 / 0.94} &
  \phantom{1}\textbf{1 / 0.75} &
  \phantom{1}\textbf{3 / 0.92} &
  \phantom{1}\textbf{5 / 1.30} &
  \phantom{1}\textbf{4 / 1.40} &
  \phantom{1}5 / 1.47 &
  22977 \\ \hline
 \multirow{2}{*}{\rotatebox{90}{{DF}}} &
  % {BPnPNet} &  \XSolidBrush
  %  &   \Checkmark
  %  &
  % - &
  % - &
  % - &
  % - &
  % - &
  %  &
  % 129/43.82 &
  % 148/51.82 &
  % 93/55.13 &
  % 261/59.06 &
  % 215/39.85 &
  % 215/43.00 &
  % 298/60.27 &
  % 302 \\
  {GoMatch~\cite{zhou2022geometry}} &  \Checkmark
   & \Checkmark
   &
  25 / 0.64 &
  283 / 8.14 &
  48 / 4.77 &
  335 / 9.94 &
  \phantom{11}48 &
   &
  \phantom{1}4 / 1.65 &
  13 / 3.86 &
  \phantom{1}9 / 5.17 &
  11 / 2.48 &
  16 / 3.32 &
  13 / 2.84 &
  89 / 21.12 &
  \phantom{11}302 \\
 &
  \textbf{DGC-GNN} & \Checkmark
   &  \Checkmark
   &
  \textbf{18 / 0.47} &
  \textbf{75 / 2.83} &
  \textbf{15 / 1.57} &
  \textbf{106 / 4.03} &
  \phantom{11}69 &
   &
  \phantom{1}\textbf{3 / 1.43} &
  \phantom{1}\textbf{5 / 1.77} &
  \phantom{1}\textbf{4 / 2.95} &
  \phantom{1}\textbf{6 / 1.61} &
  \phantom{1}\textbf{8 / 1.93} &
  \phantom{1}\textbf{8 / 2.09} &
  \textbf{71 / 19.5} &
  \phantom{11}355 \\ \hline \specialrule{2.5pt}{0.5pt}{0.5pt}
\end{tabular}
}
\caption{\textbf{ Visual Localization.} 
We report the median pose errors (cm, $^\circ$) and storage requirements (MB) on the scenes of the 7Scenes~\cite{shotton2013scene} and Cambridge-Landmarks~\cite{kendall2015posenet} datasets.
Three groups of methods are shown: end-to-end (E2E), descriptor-based (DB), and descriptor-free (DF).
We do not show BPnPNet as it fails on most scenes.
The best results are shown in bold in each group. 
} 
\vspace{-5pt}
\label{localization}
\end{center}
\end{table*}
\vspace{5pt}
\subsection{Visual Localization}
\label{Visual Localization}

Visual localization estimates the 6 degrees-of-freedom camera pose of an input query image \textit{w.r.t} a known map of the scene. 
One of the most prominent ways of approaching this problem is via establishing 2D-3D correspondences and running robust pose estimation. 
Following~\cite{zhou2022geometry}, we ran the proposed DGC-GNN to obtain matches. 
For each query image, we match its keypoints with the 3D points from the top-10 retrieved views to build the 2D-3D correspondences. 
The camera pose is then estimated by PnP-RANSAC~\cite{fischler1981random, 1217599}.  
We use two standard datasets, 7Scenes~\cite{shotton2013scene} and Cambridge Landmarks~\cite{kendall2015posenet}.  
For 7Scenes, we extract the keypoints with the SIFT detectors, and the top 10 pairs are retrieved using DenseVLAD~\cite{torii201524}. 
For Cambridge Landmarks, the keypoints are extracted by SuperPoint~\cite{detone2018superpoint} to ensure consistency with the SuperPoint-based structure-from-motion model. 
The top 10 pairs are provided by NetVLAD~\cite{Arandjelovic16}.
% \vspace{1mm}

\subsubsection{Results} 

In Table~\ref{localization}, we present the 3D model maintenance costs, privacy, storage requirements, and camera pose median errors (cm, $^\circ$) of standard descriptor-based localization techniques and descriptor-free methods. 
DGC-GNN consistently outperforms GoMatch on \textit{all} scenes by a significant margin.
On Cambridge Landmarks, the average median error of DGC-GNN is 54 cm / 2.23$^\circ$, while GoMatch leads to 173 cm / 5.87$^\circ$ error.
On 7Scenes, the average error of DGC-GNN is 15 cm / 4.47$^\circ$, and that of GoMatch is 22 cm / 5.77$^\circ$. 
DGC-GNN requires a similar amount of memory to other descriptor-free methods. 
Also, it inherits their privacy-preserving properties due to not requiring visual descriptors.

The trade-off between descriptor-based (DB) and descriptor-free (DF) algorithms is visible from the table. 
While descriptor-based ones lead to the best accuracy overall, they require excessive memory and descriptor maintenance and are susceptible to privacy attacks. 
Although the model compression method, HybridSC~\cite{camposeco2019hybrid}, shows effectiveness in storage saving, it achieves similar performance compared to DGC-GNN on the Cambridge Landmarks dataset while still requiring descriptor maintenance.
End-to-end methods (E2E) overcome these problems and achieve accurate results. 
However, their main limitation is that such approaches must be trained independently on each scene.
The proposed DGC-GNN only needs to be trained once, making it more efficient and convenient to use as an off-the-shelf tool.
\vspace{-5pt}
\subsubsection{Generalizability}
\begin{table*}[t!]
\begin{center}
\renewcommand\arraystretch{1.1}
\Huge
\resizebox{0.95\textwidth}{!}
{
\begin{tabular}{lcccclcc}
\hline \specialrule{2.5pt}{0.5pt}{0.5pt}
\multirow{2}{*}{Scenes} & \multicolumn{4}{c}{Trained on MegaDepth~\cite{li2018megadepth} (SIFT)}               &  & \multicolumn{2}{c}{Trained on ScanNet~\cite{dai2017scannet} (SIFT)} \\ \cline{2-5} \cline{7-8} 
 
                        & GoMatch (SIFT) & GoMatch (SP)  & DGC-GNN (SIFT)  & DGC-GNN (SP) &  & DGC-GNN (SIFT)       & DGC-GNN (SP)       \\ 
                        \hline \specialrule{1.5pt}{0.5pt}{0.5pt}
Chess                   & \phantom{1}4 / \phantom{1}1.65    & \phantom{1}4 / \phantom{1}1.56  & \phantom{1}\textbf{3} / \phantom{1}\textbf{1.41}    & \phantom{1}\textbf{3} / \phantom{1}1.46  &  & \phantom{1}\textbf{3} / \phantom{1}1.43            & \phantom{1}4 / \phantom{1}1.51          \\
Fire                    & 13 / \phantom{1}3.86    & 12 / \phantom{1}3.71  & \phantom{1}\textbf{5} / \phantom{1}1.81    & \phantom{1}7 / \phantom{1}2.30  &  & \phantom{1}\textbf{5} / \phantom{1}\textbf{1.77}            & \phantom{1}6 / \phantom{1}2.03          \\
Heads                   & \phantom{1}9 / \phantom{1}5.17    & \phantom{1}5 / \phantom{1}3.43  & \phantom{1}5 / \phantom{1}3.13    & \phantom{1}\textbf{4} / \phantom{1}\textbf{2.78}  &  & \phantom{1}\textbf{4} / \phantom{1}2.95            & \phantom{1}\textbf{4} / \phantom{1}3.02          \\
Office                  & 11 / \phantom{1}2.48    & \phantom{1}7 / \phantom{1}1.76  & \phantom{1}7 / \phantom{1}1.66    & \phantom{1}7 / \phantom{1}1.66  &  & \phantom{1}\textbf{6} / \phantom{1}\textbf{1.61}            & \phantom{1}7 / \phantom{1}1.66          \\
Pumpkin                 & 16 / \phantom{1}3.32    & 28 / \phantom{1}5.65  & \phantom{1}\textbf{8} / \phantom{1}2.03    & 12 / \phantom{1}2.75  &  & \phantom{1}\textbf{8} / \phantom{1}\textbf{1.93}            & 10 / \phantom{1}2.38          \\
Redkitchen              & 13 / \phantom{1}2.84    & 14 / \phantom{1}3.03  & \phantom{1}\textbf{8} / \phantom{1}2.14    & 10 / \phantom{1}2.36  &  & \phantom{1}\textbf{8} / \phantom{1}\textbf{2.09}            & \phantom{1}9 / \phantom{1}2.28          \\
Stairs                  & 89 / 21.12   & 58 / 13.12 & 83 / 21.53   & \textbf{55 / 13.05} &  & 71 / 19.50            & 58 / 14.32         \\ \hline
All & 22 / \phantom{1}5.78 & 18 / \phantom{1}4.61 & 17 / \phantom{1}4.82 & \textbf{14} / \phantom{1}\textbf{3.77} & & 15 / \phantom{1}4.47 & \textbf{14} / \phantom{1}3.89 \\ \hline \specialrule{2.5pt}{0.5pt}{0.5pt}
\end{tabular}

}
\caption{\textbf{Model Generalizability on Visual Localization Task.} We report the translation and rotation median error (cm / $^\circ$) on 7Scenes dataset~\cite{shotton2013scene}. We evaluate the models trained on MegaDepth~\cite{li2018megadepth} and ScanNet~\cite{dai2017scannet} datasets. The best performance is bold.
} 
\vspace{-10pt}
\label{tab:gene}
\end{center}
\end{table*}
Similar to~\cite{zhou2022geometry}, we discuss the generalizability of our DGC-GNN model on the visual localization task across different training and evaluation scenes. 
Specifically, we investigate the performance of our model when trained on MegaDepth~\cite{li2018megadepth} and ScanNet~\cite{dai2017scannet} and evaluated on the 7Scenes dataset~\cite{shotton2013scene}. 
We also explore the impact of using different keypoint detectors, namely SIFT~\cite{lowe2004distinctive} and SuperPoint~\cite{detone2018superpoint}, during the evaluation.

These experiments are summarized in Table~\ref{tab:gene}, providing an overview of the performance of DGC-GNN under different training and evaluation conditions. 
While the best overall performance is achieved by training on the MegaDepth dataset with SIFT features, the results are similar in both training scenarios, showcasing that the proposed method generalizes well to unseen data. 
While we train on SIFT features, the best results are achieved by using SuperPoint features at inference time.
This demonstrates that DGC-GNN is insensitive to the features used and can be utilized off-the-shelf even without retraining to our specific scenario. 

\section{Conclusion}
In conclusion, this paper introduces DGC-GNN, a novel graph-based pipeline for visual descriptor-free 2D-3D matching that effectively leverages geometric and color cues in a global-to-local manner. 
Our global-to-local procedure encodes both Euclidean and angular relations at a coarse level, forming a geometric embedding to guide the local point matching. 
By employing a cluster-based transformer, we enable efficient information passing within local clusters, ultimately leading to significant improvements in the number of correct matches and the accuracy of pose estimation.
Compared to the state-of-the-art descriptor-free matcher GoMatch~\cite{zhou2022geometry}, the proposed DGC-GNN demonstrates a substantial improvement, doubling the accuracy on real-world and large-scale datasets. 
Furthermore, it results in significantly increased localization accuracy. 
These advancements contribute to reducing the gap between descriptor-based and descriptor-free methods while addressing the limitations of descriptor-based ones, such as memory footprint, maintenance costs, and susceptibility to privacy attacks.

\noindent
\textbf{Limitations.} The primary limitation of our proposed DGC-GNN method lies in its performance being inferior to traditional descriptor-based algorithms. 
The performance difference can be attributed to the insufficiency of unique 3D structures in the geometry, which hinders the ability of the algorithm to identify distinct matches in real-world scenarios. 
Although DGC-GNN demonstrates a notable improvement over existing descriptor-free approaches, there remains a performance gap to overcome in order to achieve results on par with or superior to those of descriptor-based methods.

\noindent
\textbf{Acknowledgements.} This work was supported by the Academy of Finland (grants No. 327911, No. 353138) and the Hasler Stiftung Research Grant via the ETH Zurich Foundation. We acknowledge the computational resources provided by the CSC-IT Center for Science, Finland.

{
    \small
    \bibliographystyle{ieeenat_fullname}
    \bibliography{cvpr}
}

\clearpage
% \section{Supplementary Material}

\appendix

\maketitlesupplementary

\section{Training and Evaluation Details}
\noindent
\textbf{Dataset Generation.} 
The training data generation process for MegaDepth~\cite{li2018megadepth} follows the methodology outlined in GoMatch~\cite{zhou2022geometry}. 
The undistorted SfM model reconstructions used in MegaDepth are provided by D2Net~\cite{dusmanu2019d2}. For training, we sample up to 500 images from each scene. For each sampled image, we select the top-$k$ co-visible views that have at least 35\% image overlap. This ensures that there are enough matches for training.  The overlapping score is computed by dividing the number of co-visible 3D points by the total number of points in the training image.

In the case of ScanNet~\cite{dai2017scannet}, a similar procedure is conducted. 
We also sample up to 500 images from each scene for the training set generation. The co-visible images are obtained using the co-visible scores provided by LoFTR~\cite{sun2021loftr}. We extract all the co-visible views of a training image with co-visible scores larger than 0.65. 
Then, we randomly sample the top-$k$ views for training. 
Since ScanNet is an RGB-D dataset without an SfM reconstruction, we obtain the 3D points for each image by projecting the detected 2D keypoints with valid depth to 3D. 
By doing this for each image, we reconstruct a sparse 3D point cloud based on the detected 2D keypoints. Note that the correspondence between different co-visible frames is not required in this case.

In total, for MegaDepth, we generate a training set consisting of 25,624 images from 99 scenes and a test set comprising 12,399 images covering 53 scenes. 
For ScanNet, we create a training set with 52,008 images from 105 scenes. The test set for ScanNet consists of 14,892 query images from 30 scenes. The data generation of 7Scenes~\cite{shotton2013scene} and Cambridge dataset~\cite{kendall2015posenet} follows the same procedure in ~\cite{zhou2022geometry}.

\begin{figure}[t]
  \centering
  \includegraphics[width=0.99\linewidth]{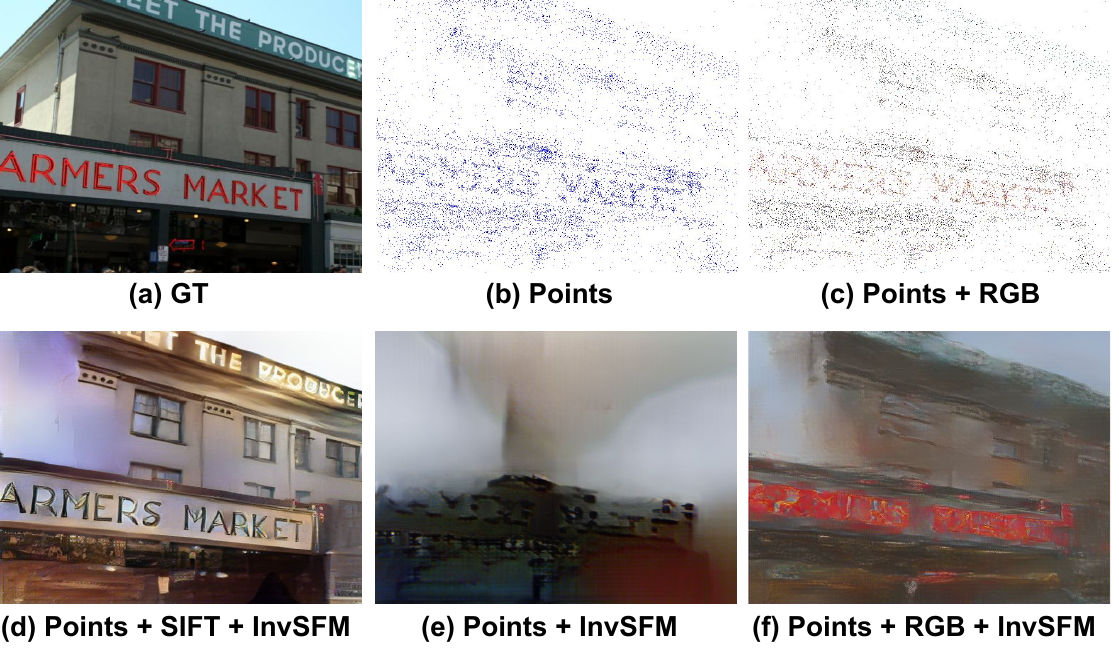}
  \vspace{-10pt}
  \caption{\textbf{Points Reprojection and Image Recovery Example}. 
}
  \label{fig: rebuttal}
\end{figure}

\begin{table}
\centering
\renewcommand\arraystretch{1.4}
\resizebox{.46\textwidth}{!}
{
\begin{tabular}{lcclccc}
\hline
\specialrule{1.5pt}{0.5pt}{0.5pt}
\multirow{2}{*}{Method} & \multicolumn{2}{c}{Reprojection}            &  & \multicolumn{3}{c}{InvSFM~\cite{pittaluga2019revealing}}                                         \\ \cline{2-3} \cline{5-7} 
                        & Points               & Points+RGB           &  & Points               & Points+RGB           & Points+SIFT          \\ \hline
SSIM ($\downarrow$)                    & \textbf{0.240}                & 0.258                &  & \textbf{0.352}                & 0.375                & 0.476                \\
\hline
\specialrule{1.5pt}{0.5pt}{0.5pt}
\end{tabular}
}
\caption{\textbf{SSIM Results.} We evaluate the SSIM from Point Reprojection and Image Recovering, adding RGB to points leads only to a slight SSIM increase on both reprojection and image recovery.} 
\label{diff_GT}
\end{table}

\noindent
\textbf{Inference.} 
We consider a query with at least 10 keypoints as valid input. The 3D points from the top-$k$ retrieved database images are then applied to match against the queries with our proposed pipeline. We use the Sinkhorn algorithm~\cite{sinkhorn1967concerning, cuturi2013sinkhorn} to optimize the extended cost matrix $\bar{\mathcal{M}} \in \mathbb{R}^{N+1, M+1}$ in an iterative manner with up to 20 iterations to obtain the initial matches. The final matches are obtained by filtering the matches with matching confidence $\theta < 0.5$ in the outlier rejection module. 
For the visual localization task, the camera poses are estimated by the P3P solver with RANSAC~\cite{fischler1981random} implemented in OpenCV~\cite{bradski2000opencv} and then refined by Levenberg-Marquardt~\cite{more2006levenberg} algorithm on the inliers matches, minimizing the reprojection error. 

\begin{figure*}[th]
  \centering
  \includegraphics[width=0.95\linewidth]{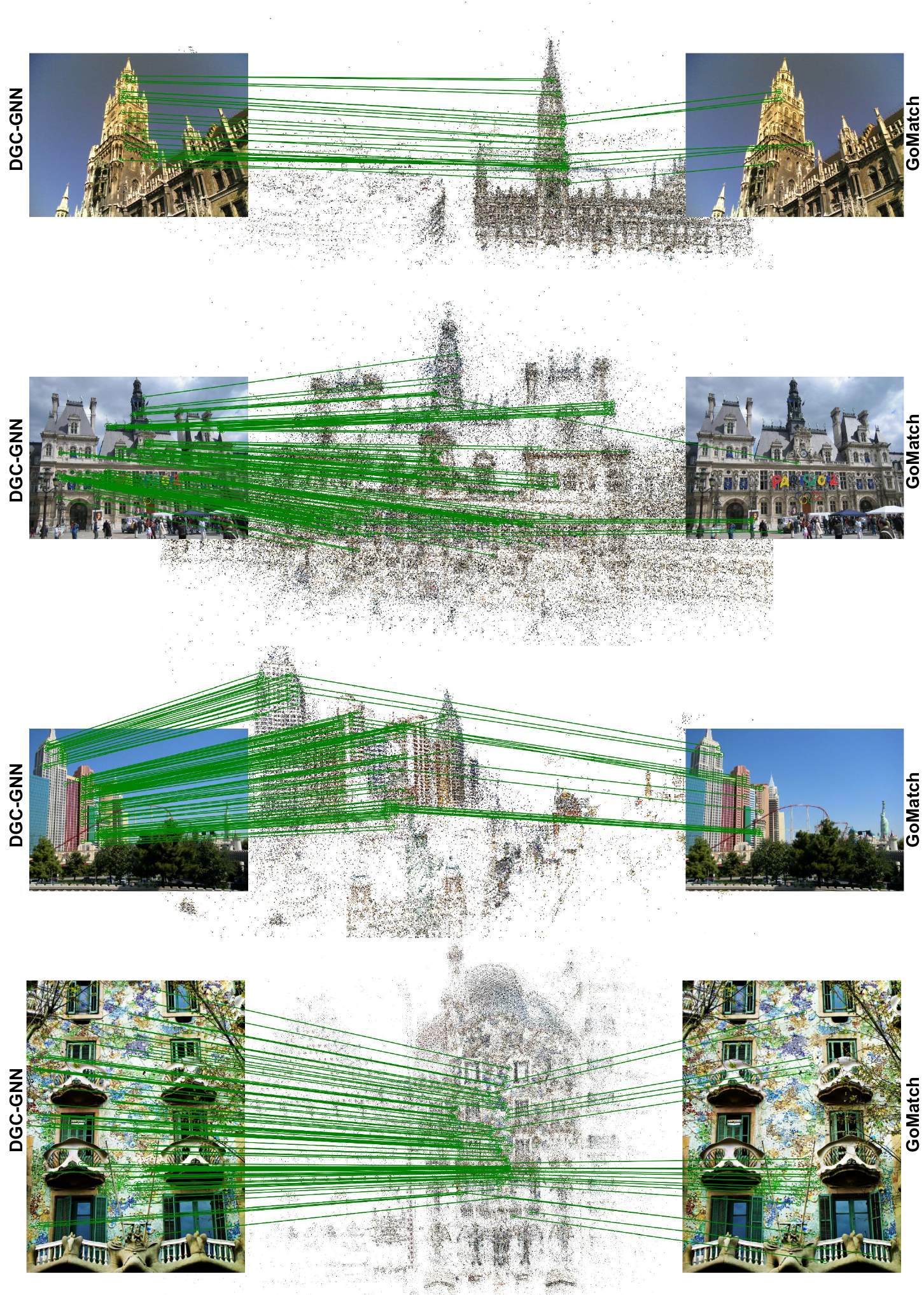}
  \caption{\textbf{2D-3D Matching} (shown by green lines) with the proposed DGC-GNN and GoMatch~\cite{zhou2022geometry}. 
}
  \label{fig: viz}
\end{figure*}

\section{Privacy Issue of RGB Points}
We investigate the impact on privacy resulting from incorporating RGB information into pixels and points. To assess this, we compute the Structural Similarity Index Measure (SSIM) for 3D points reprojected onto the image plane against the ground truth (GT) images on MegaDepth over 500 images from multiple scenes. Additionally, we recover the images from points + RGB and points + descriptors with InvSFM~\cite{pittaluga2019revealing} to calculate the SSIM against the GTs. The findings are detailed in Table~\ref{diff_GT} and Fig~\ref{fig: rebuttal}. The addition of RGB data to points results in only a marginal increase in SSIM for both direct reprojection and image reconstruction via InvSFM, significantly less than what is achieved by incorporating SIFT descriptors. It is worth noting that denser point clouds might provide sufficient context, potentially leading to privacy concerns. However, in our setting, we mitigate this risk by limiting the number of keypoints from each database image to a maximum of 1024.

\section{Additional Results}
\noindent
\textbf{Qualitative Results.} More visualizations of inlier matches provided by DGC-GNN and GoMatch on MegaDepth are shown in Fig.~\ref{fig: viz}. DGC-GNN consistently finds more correct matches on multiple scenes, highlighting the effectiveness of the proposed method.
\begin{table*}[t!]
\begin{center}
\renewcommand\arraystretch{1.4}
\setlength{\tabcolsep}{3pt}
\resizebox{.99\textwidth}{!}
{
\begin{tabular}{lcccccccc}
\hline 
\specialrule{1.5pt}{0.5pt}{0.5pt}
\multirow{2}{*}{Methods} &
  \multirow{2}{*}{Global} &
  \multirow{2}{*}{C. Att.} &
  \multirow{2}{*}{Color} &
  \multirow{2}{*}{Ang.} &
  \multirow{2}{*}{Cluster} &
  Reproj. AUC (\%) &
  Rotation ($^\circ$) &
  Translation \\
& & & & &  &@1 / 5 / 10px  ($\uparrow$)           & \multicolumn{2}{c}{Quantile@25 / 50 / 75\% ($\downarrow$)}        \\ 
\specialrule{1.5pt}{0.5pt}{0.5pt}
GoMatch~\cite{zhou2022geometry} (w/o OR)                   & & & & & &
\phantom{1}4.47 / 17.95 / 23.42 &	1.29 / 11.85 / 33.60& 0.11 / 1.18 / 3.58 \\
GoMatch~\cite{zhou2022geometry}                   & & & & & &
\phantom{1}5.67 / 22.43 / 28.01 & 0.60 / 10.08 / 34.63 & 0.06 / 1.06 / 3.73 \\
\hline

\multirow{5}{*}{Variants}  & G.Emb & & &    & K-means     & \phantom{1}7.68 / 28.41 / 34.36            &  0.28 / \phantom{1}6.78 / 34.52           & 0.03 / 0.73 / 3.77      \\
& G.Label & \Checkmark & &    & K-means     & \phantom{1}7.13 / 27.33 / 33.18           & 0.31 / \phantom{1}7.34 / 33.63           & 0.03 / 0.76  / 3.64      \\
                                        & G.Emb & \Checkmark &  &     & K-means& \phantom{1}8.10 / 30.64 / 37.07           &   0.24 / \phantom{1}4.48 / 34.30         &  0.03 / 0.63 / 3.51      \\
                                        & G.Emb & \Checkmark & \Checkmark &      & K-means&  \phantom{1}9.82 / 35.29 / 41.16           &   0.17  / \phantom{1}2.88 / 31.74         &  0.02 / 0.27 / 3.24      \\
                                        & G.Emb & \Checkmark & \Checkmark & \Checkmark & Mean-shift &  10.07 / 36.01 / 43.03          &  0.16  / \phantom{1}2.15 / 28.99           &  \textbf{0.01} / 0.20 / 3.26       \\ \hline
DGC-GNN (w/o OR)                & G.Emb & \Checkmark & \Checkmark & \Checkmark & K-means &  \phantom{1}8.56 / 30.79 / 37.03 & 0.22 / \phantom{1}4.85 / 30.07 & 0.02 / 0.47 / 3.10 \\ 
\textbf{DGC-GNN}                  & G.Emb & \Checkmark & \Checkmark & \Checkmark & K-means &  \textbf{10.20} / \textbf{37.64} / \textbf{44.04 } & \textbf{0.15 /  \phantom{1}1.53 /  27.93} & \textbf{0.01 / 0.15 / 3.00} \\ 
\specialrule{1.5pt}{0.5pt}{0.5pt}
\end{tabular}
}
\caption{\textbf{Additional Ablation Results.}
AUC scores thresholded at 1, 5, and 10 pixels on $k=1$; rotation and translation error quantiles at 25, 50, 75\% with the proposed components added one by one to the GoMatch pipeline. 
} 
\label{tab:abla_supp}
\end{center}
\end{table*}
\begin{figure*}[t]
  \centering
  \begin{subfigure}{0.49\textwidth}
    \centering
    \includegraphics[width=\textwidth]{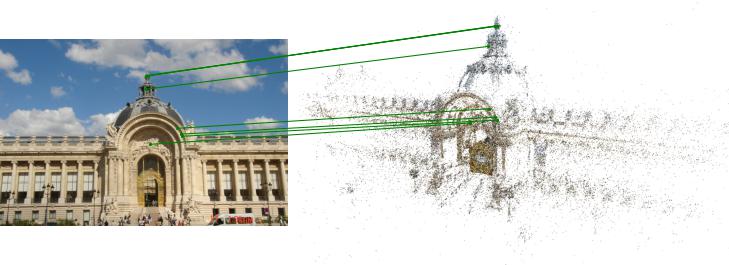}
    \caption{GoMatch}
    \label{fig:subfig1}
  \end{subfigure}
  \hfill
  \begin{subfigure}{0.49\textwidth}
    \centering
    \includegraphics[width=\textwidth]{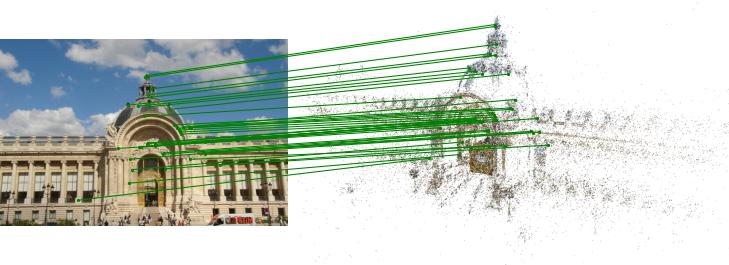}
    \caption{+G. Emb, +Cluster Attention}
    \label{fig:subfig2}
  \end{subfigure}

  \begin{subfigure}{0.49\textwidth}
    \centering
    \includegraphics[width=\textwidth]{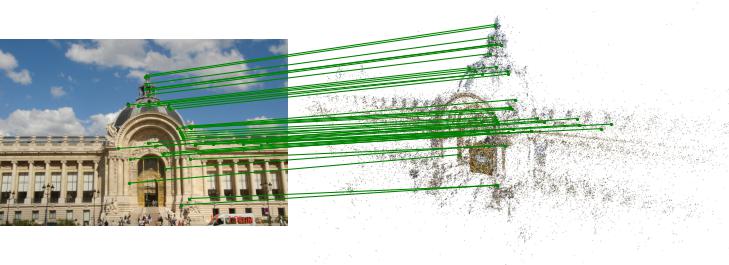}
    \caption{+G. Emb, +Cluster Attention, +Color}
    \label{fig:subfig3}
  \end{subfigure}
  \hfill
  \begin{subfigure}{0.49\textwidth}
    \centering
    \includegraphics[width=\textwidth]{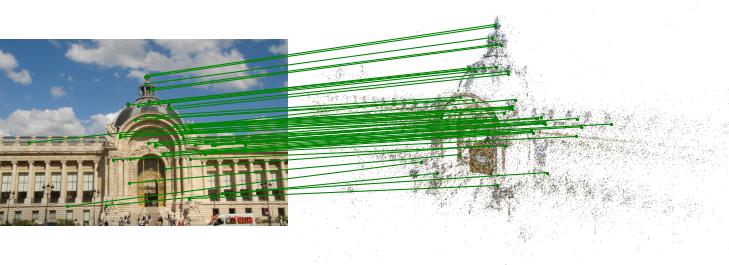}
    \caption{DGC-GNN}
    \label{fig:subfig4}
  \end{subfigure}
  \caption{
  \textbf{Qualitative Matching Results of Different Architectures.} We visualize the number of inlier matches after the PnP-RANSAC with different architectures (shown by green lines ).
  } 
  \label{fig:subfigures}
\end{figure*}

\noindent
\textbf{Additional Ablation Results.} In addition to the ablation results presented in the main paper, we also provide ablation results for single-view matching with $k=1$ on MegaDepth~\cite{li2018megadepth}. Furthermore, we conduct two additional ablations to investigate the impact of different component selections. Firstly, we compare the effectiveness of the geometric global embedding (G. Emb.) used in the main paper with the global clustering label embedding (G. Label). Instead of encoding geometric cues, we encode the label of each global cluster and concatenate it to the local point feature. Then, we explore the selection of different clustering algorithms. We compare the performance of K-Means and Mean-Shift clustering algorithms in our pipeline. Last, we study the effectiveness of the outlier rejection (OR) network.

The results are presented in Table~\ref{tab:abla_supp}. We observe similar conclusions for each component as in the main paper. The results obtained using the global label embedding (G. Label) with cluster attention (C.Att) show even worse performance compared to geometric embedding (G. Emb.) only, indicating the superiority of our clustering-based geometric embedding over the label embedding and highlighting the importance of incorporating geometric cues in the embedding process for effective point matching. Regarding the impact of different clustering algorithms, we only observe a minor difference in K-Means and Mean-Shift results, suggesting that our approach is robust to the choice of the clustering algorithm. The results also demonstrate that outlier rejection is an essential post-processing module to achieve good performance.
In addition to the numerical results, we visualize the inlier matches (see Fig.~\ref{fig:subfigures}) to provide deep insights into the behavior and performance of different architectures.

\noindent
\textbf{Hyperparameters analysis.} 
Besides the component ablations, we also give an in-depth analysis of the hyperparameters used in our main pipeline. 
Here, we add additional ablations on the number of input keypoints, the number of cluster groups at the coarse level, the number of nearest neighbors in the local graph build, and the outlier rejection threshold by retraining our DGC-GNN. 
The results are presented in Table~\ref{tab:hyparam_abla}. We observe that DGC-GNN with G.\ Clusters = 10 and Local NN = 10 achieves overall the best performance. 
Setting the outlier rejection threshold to 0.7 leads to the best performance. 
However, the results are stable across different configurations, indicating robustness to the parameter setting. 

\begin{table*}[t!]
\begin{center}
\renewcommand\arraystretch{1.3}
\setlength{\tabcolsep}{3pt}
\resizebox{.99\textwidth}{!}
{
\begin{tabular}{lcccccc}
\hline
\specialrule{1.5pt}{0.5pt}{0.5pt}
\multirow{2}{*}{Methods}     & \multirow{2}{*}{G. Cluters} & \multirow{2}{*}{Local NN} & \multirow{2}{*}{OR Threshold} & Reproj. AUC (\%)                  & Rotation ($^\circ$)                    & Translation                 \\
                             &                             &                           &                               & @1 / 5 / 10px  ($\uparrow$)    & \multicolumn{2}{c}{Quantile@25 / 50 / 75\% ($\downarrow$)} \\ \specialrule{1.5pt}{0.5pt}{0.5pt}
DGC-GNN                      & 10                          & 10                         & 0.5                           & 15.30 / 51.70 / 60.01          & \textbf{0.07} / 0.26 / \phantom{1}5.41           & \textbf{0.01} / \textbf{0.02} / 0.57    \\ \hline
\multirow{7}{*}{HyperParam.} & 5                           & 10                       & 0.5                           & 14.73 / 50.12 / 58.26          & 0.08 / 0.28 / \phantom{1}8.76           & \textbf{0.01} / 0.03 / 0.99          \\
                             & 15                          & 10                        & 0.5                           & 15.14 / 50.56 / 58.62          & \textbf{0.07} / 0.28 / \phantom{1}7.66             & \textbf{0.01} / 0.03 / 0.89          \\ \cline{2-7} 
                             & 10                          & 20                        & 0.5                           &              14.77 / 49.84 / 57.97                  &                 \textbf{0.07} / 0.29 / \phantom{1}8.26            &           \textbf{0.01} / 0.03 / 0.90   \\
                                                          & 10                          & 30                         & 0.5                           &        14.75 / 50.95 / 59.45	                       &            0.08 / 0.28 / \phantom{1}5.48                 &   \textbf{0.01} / 0.03 / 0.58  
                             \\ \cline{2-7} 
                             % & 10                          & 10                         & 0.1                           & 10.53 / 39.04 / 47.77          & 0.12 / 0.99 / 14.05          & \textbf{0.01} / 0.09 / 1.55          \\       
                             & 10                          & 10                         & 0.3                           & 13.28 / 46.44 / 55.05          & 0.08 / 0.43 / \phantom{1}8.63           & \textbf{0.01} / 0.04 / 0.98          \\
                             & 10                          & 10                         & 0.7                           & \textbf{16.63 / 56.26 / 64.46} & \textbf{0.07 / 0.19 / \phantom{1}2.58}  & \textbf{0.01 / \textbf{0.02} / 0.27}

        \\ \specialrule{1.5pt}{0.5pt}{0.5pt}
\end{tabular}
}
\vspace{10pt}
\caption{\textbf{Ablation Study on Hyperparameters.}
We report the results of ablations with retrieved image $k=10$ on the number of global clusters, the number of nearest neighbour points for local graph build, and different thresholds for outlier rejection. The best results are bold.
} 
\label{tab:hyparam_abla}
\end{center}
\end{table*}

\noindent
\textbf{Matching Results in pixel threshold.} 
As mentioned in the main paper, we selected the ground truth matches in normalized image coordinates. 
The described GT difference only affects the reprojection AUC scores. Here, we present the matching results in Table~\ref{diff_GT} by selecting the ground truth matches in pixel coordinate with $1$ pixel threshold as done in~\cite{zhou2022geometry}. Our conclusions still hold.

\section{Model Parameters and Timing}
We discuss the model parameters and running time of DGC-GNN in this section. DGC-GNN incorporates global geometric embedding and local clustering attention, which has around 5.7 million trainable parameters and an estimated model size of 22.6 MB. The average inference time for each image pair over the Megadepth evaluation queries is 77.8ms. It roughly breaks down into point encoding (24 ms), global geometric embedding (14 ms), cluster-based attention (22 ms), optimal transport (7 ms), and outlier rejection (8 ms). The measurements are conducted on a 32GB NVIDIA Telsa V100 GPU with a maximum of 1024 keypoints.

\end{document}